# Unsupervised Domain Adaptation Via Data Pruning

Andrea Napoli, Paul White
*Institute of Sound and Vibration Research*
*University of Southampton, UK*
{an1g18, P.R.White}@soton.ac.uk

*Abstract*—The removal of carefully-selected examples from training data has recently emerged as an effective way of improving the robustness of machine learning models. However, the best way to select these examples remains an open question. In this paper, we consider the problem from the perspective of unsupervised domain adaptation (UDA). We propose AdaPrune, a method for UDA whereby training examples are removed to attempt to align the training distribution to that of the target data. By adopting the maximum mean discrepancy (MMD) as the criterion for alignment, the problem can be neatly formulated and solved as an integer quadratic program. We evaluate our approach on a real-world domain shift task of bioacoustic event detection. As a method for UDA, we show that AdaPrune outperforms related techniques, and is complementary to other UDA algorithms such as CORAL. Our analysis of the relationship between the MMD and model accuracy, along with t-SNE plots, validate the proposed method as a principled and well-founded way of performing data pruning.

## I. Introduction

The development of *robust* models has been a longstanding goal in machine learning, and is crucial to the widespread deployment of AI. The issue has spurred significant innovation, and has been approached from a wide variety of angles. Yet, a range of recent work has shown, from both empirical and theoretical standpoints, that many of the methods proposed are not any better in practice than vanilla training by empirical risk minimisation (ERM) [1–4].

At the same time, multiple analyses have shown the single most important factor which determines a model's robustness to be the *composition of the training set*, with variables such as the model architecture, learning process, loss function, hyperparameters or training set size having little to no effect [5–7].

As a result, data pruning (also referred to as data filtering or coreset selection) has emerged as a popular approach to improving the robustness properties of the training dataset [8–10]. Pruning has a well-established literature in contexts such as outlier removal [11] or the reduction of memory or annotation requirements [12–14], and many selection criteria exist. The reliability of these criteria, however, has at times been found to be poor [15].

We note that distributional robustness is defined *relative to the test distribution* – that is, it is the *relation* between the training and test distributions which determines robustness [8]. Therefore, we posit that access to the test distribution is required for the pruning algorithm to be able to make an informed choice – and, consequently, that this problem is most suitably considered from the perspective of unsupervised domain adaptation (UDA).

UDA is a well-established paradigm which attempts to increase robustness to domain shifts by aligning the training and test distributions – that is, by minimising some measure of the statistical distance between them. In the context of pruning, distribution alignment must be balanced with the need to not reduce the training set so far that the model overfits and loses its generalisation capability. Additionally, as pruning is a discrete decision problem, the choice of distance measure should be optimisable as such. Recently, an adaptive pruning strategy based on a domain discriminator network was proposed for knowledge distillation in [16].

In the following sections, we formalise data pruning as a UDA problem. Then, we show that adopting the maximum mean discrepancy (MMD) as the alignment criterion results in an integer quadratic programming (IQP) formulation which can be readily solved using standard optimisation software. Finally, we conduct a substantive analysis on a cross-dataset bioacoustic event detection task, where we find AdaPrune achieves higher out-of-distribution performance than related UDA algorithms.

### A. Data Pruning as a UDA Problem

Given labelled examples $\mathcal{D}_s = \{(x_i^s, y_i^s)\}_{i=1}^{N_s}$ from a source distribution $P_s(x, y)$, and unlabelled examples $\mathcal{D}_t = \{x_j^t\}_{j=1}^{N_t}$ from a different target distribution $P_t(x, y)$, the goal of UDA is to produce a model $\Theta : \mathcal{X} \to \mathcal{Y}$ such that the risk under $P_t$

$$R_t(\Theta) \triangleq \mathbb{E}_{(x,y) \sim P_t}[L(\Theta(x), y)] \tag{1}$$

is minimised. For AdaPrune, we propose to fit $\Theta$ to a subset $\mathcal{D}_{ss} \subseteq \mathcal{D}_s$, i.e., to minimise an empirical proxy to $R_t$, given by

$$R_{ss}(\Theta) = \frac{1}{N_{ss}} \sum_{i=1}^{N_s} u_i L(\Theta(x_i^s), y_i^s), \tag{2}$$

where $u_i \in \{0,1\}$ is a binary variable indicating whether training pair $(x_i^s, y_i^s)$ is included in $\mathcal{D}_{ss}$, and $N_{ss} = |\mathcal{D}_{ss}|$ is a hyperparameter to be determined. Then, the "adaptation" reduces to selecting appropriate values for $u_i$. Note, as with the related technique of importance weighting, this formulation assumes that the support of $P_t$ is contained by that of $P_s$ [17].

Let $u = [u_1, ..., u_{N_s}]^T$. Our objective is to select $u$ to minimise the discrepancy between $\mathcal{D}_{ss}$ and $\mathcal{D}_t$, which we measure using the MMD. Given the potentially high dimensionality of $\mathcal{X}$, doing this directly in the input space can be impractical. Therefore, we propose instead to first extract feature embeddings $z_i^s = \Theta_F(x_i^s)$ and $z_j^t = $ $\Theta_F(x_j^t)$ respectively, where $\Theta_F : \mathcal{X} \to \mathcal{Z}$ is a featuriser pre-trained on the entirety of $\mathcal{D}_s$.

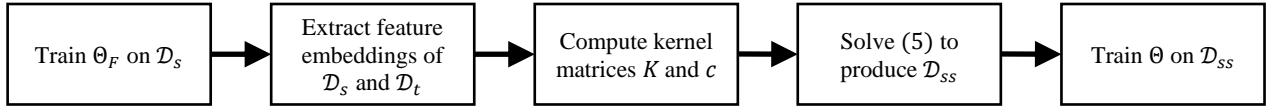

Figure 1: AdaPrune, unsupervised domain adaptation via data pruning.

The (squared) empirical MMD between the feature embeddings of $\mathcal{D}_{ss}$ and $\mathcal{D}_t$ is given by

$$\text{MMD}^2 = \left\| \frac{1}{N_{ss}} \sum_{i=1}^{N_s} u_i \phi(z_i^s) - \frac{1}{N_t} \sum_{j=1}^{N_t} \phi(z_j^t) \right\|_{\mathcal{H}}^2, \quad (3)$$

where $\mathcal{H}$ is a (potentially infinite-dimensional) reproducing kernel Hilbert space, and $\phi : \mathcal{Z} \to \mathcal{H}$ is an implicit mapping. $\mathcal{H}$ is associated with a unique positive-definite kernel $\kappa : \mathcal{Z} \times \mathcal{Z} \to \mathbb{R}$ for which the reproducing property $\kappa(z_i, z_j) = \langle \phi(z_i), \phi(z_j) \rangle_{\mathcal{H}}$ is satisfied. This property allows us to adopt the kernel trick, whereby inner products (and thus norms) in $\mathcal{H}$ are substituted by evaluations of $\kappa$. In this case, (3) becomes

$$\text{MMD}^2 = \frac{1}{N_{ss}^2} \sum_{i=1}^{N_s} \sum_{j=1}^{N_s} u_i u_j \kappa(z_i^s, z_j^s) + \frac{1}{N_t^2} \sum_{i=1}^{N_t} \sum_{j=1}^{N_t} \kappa(z_i^t, z_j^t)$$
$$- \frac{2}{N_{ss} N_t} \sum_{i=1}^{N_s} u_i \sum_{j=1}^{N_t} \kappa(z_i^s, z_j^t). \quad (4)$$

Define $K \in \mathbb{R}^{N_s \times N_s}$, $K_{ij} = \kappa(z_i^s, z_j^s)$ and $c \in \mathbb{R}^{N_s}$, $c_i = \sum_{j=1}^{N_t} \kappa(z_i^s, z_j^t)$. The minimisation problem can now be formulated as the binary quadratic program

$$\min_u \frac{1}{N_{ss}} u^T K u - \frac{2}{N_t} c^T u \quad (5)$$

subject to $u_i \in \{0,1\}$ and $\sum_{i=1}^{N_s} u_i = N_{ss}$. For $\kappa$, we adopt a radial basis function (RBF) mixture kernel [18], given by

$$\kappa(z_i, z_j) = \sum_{\gamma \in \mathcal{G}} e^{-\gamma \|z_i - z_j\|^2} \quad (7)$$

with $\mathcal{G} = \{0.001, 0.01, 0.1, 1, 10\}$. The RBF allows us to capture nonlinear relationships between domains, whilst mixing RBFs with different bandwidths provides an efficient solution to multi-scale analysis. The overall training pipeline is shown in Figure 1.

### B. Relation to Kernel Mean Matching

The above formulation is closely related to the UDA method of kernel mean matching (KMM) [17], in that both methods seek to minimise the MMD between the training and test data by way of adjusting the training distribution. However, KMM achieves this by re-weighting, rather than removing, training instances. It has been shown that reweighting algorithms (of which classical importance weighting [17] and Distributionally Robust Optimisation [19] are other examples) are fundamentally limited by the implicit bias of gradient descent, and ultimately tend to converge to the same solutions as ERM – so cannot generalise any better [2, 20].

The result suggests that down-weighting irrelevant examples is insufficient: these should instead be removed entirely from the training data. Thus, enforcing this stronger constraint explicitly, as we propose in our formulation, should result in a more generalisable solution.

### C. Relation to landmark selection

Our method is also closely related to landmark selection [21–23], in which a subset of training data most similar to the test data is selected. Again, there are key differences in both how the subset is selected and how it is used. Firstly, the landmarks are not used to adjust the training distribution, but to augment the target data to facilitate UDA downstream. When optimising (5), AdaPrune explicitly imposes the binary constraint on $u$, which the solver then enforces via sophisticated branch-and-bound heuristics. In contrast, [21] solves a fractional relaxation and then thresholds the result, giving a lower quality solution. AdaPrune also constrains the subset size; this is then tuned separately as a hyperparameter at the model selection stage. This satisfies the important requirement made in the previous section, as it allows the balance between distribution alignment and in-distribution accuracy to be properly controlled, thus greatly reducing the risk of over-pruning.

## II. EXPERIMENTS

In this section, we evaluate the proposed method on a real-world domain shift problem, namely, the detection of humpback whale calls across data from different acoustic monitoring programs [24]. We note that bioacoustic monitoring is a particularly suitable application on which to apply data pruning. Given all the factors which can vary across monitoring programs (such as recording equipment, environmental conditions, the distribution of nontarget sound events, or the call repertoire of the animals themselves), AdaPrune provides a means to say which training data is most relevant to build a bespoke model for a newly-collected set of data, whilst discarding bad examples which may cause negative transfer [25]. Moreover, we emphasise that no additional meta-data (or labels) from the new location are required.

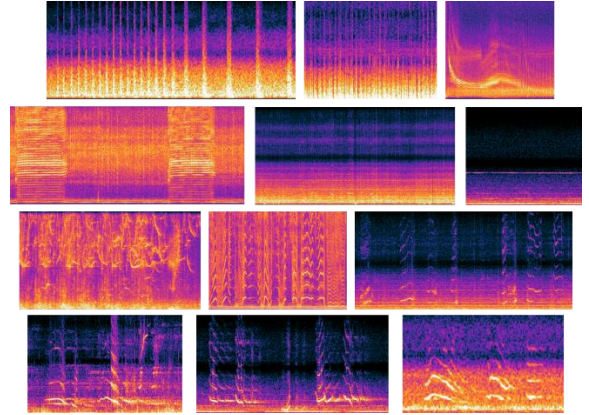

Figure 2: Some exemplar spectrograms of sounds in the dataset (5 kHz bandwidth, time axis scales variable). Top row: sperm whale clicks, pilot whale clicks, seal vocalisations. Second row: minke whale boings, right whale calls in strong vessel noise, electrical interference. Third row: dolphin whistles, dolphin creaks, right whale calls. Bottom row: three humpback whale calls.

Table 1: Test domain accuracy (%) for each training algorithm.

| Method | Test accuracy by held-out domain (%) | | | | |
|---|---|---|---|---|---|
| | 1 | 2 | 3 | 4 | Average |
| ERM | 67.9 ± 2.5 | 76.2 ± 5.8 | 67.8 ± 4.0 | 94.3 ± 0.4 | 76.5 ± 2.6 |
| + Landmarks | **77.0 ± 1.9** | 77.4 ± 5.6 | 66.7 ± 1.1 | 49.1 ± 0.8 | 67.5 ± 1.4 |
| + KMM | 68.4 ± 7.6 | 73.0 ± 1.5 | **82.4 ± 5.7** | **94.9 ± 0.3** | 79.7 ± 3.7 |
| + AdaPrune | 71.9 ± 3.0 | **77.5 ± 3.0** | 80.5 ± 6.1 | 93.8 ± 0.3 | **80.9 ± 2.9** |
| CORAL | 81.8 ± 1.9 | 91.6 ± 4.0 | 78.1 ± 3.6 | 94.8 ± 1.5 | 86.6 ± 1.9 |
| + Landmarks | 80.0 ± 3.1 | 80.8 ± 6.5 | 80.4 ± 7.4 | 85.0 ± 4.3 | 81.5 ± 4.0 |
| + KMM | 60.7 ± 2.8 | 90.8 ± 4.2 | 78.8 ± 7.0 | **95.2 ± 0.8** | 81.4 ± 1.4 |
| + AdaPrune | **77.5 ± 5.4** | **97.8 ± 1.0** | **93.0 ± 2.3** | 93.1 ± 1.7 | **90.4 ± 2.2** |

The dataset comprises 8,000 samples of underwater sound events split equally across 4 recording locations (Madagascar, UK, Hawaii, and Australia). Each sample is a PCEN-normalised [26] mel-spectrogram of a 4-second audio clip sampled at 10 kHz, labelled as either "humpback whale" or "not humpback whale". Some exemplar spectrograms are shown in Figure 2 (note, these images are linearly frequency scaled and pre-PCEN).

Experiments are conducted using the DomainBed framework [1], with the Gurobi Optimizer [27] used to solve the IQP. The model comprises a simple 4-layer CNN architecture, with 16 filters per layer and RELU activations. This is trained on 3 locations ("domains") at a time, using the Adam optimiser, for 2,000 iterations. The remaining location is split into a "UDA set" (referred to as $\mathcal{D}_t$ above), used for adaptation, and an independent test set, used for evaluation.

Since no relevant pre-trained feature extractor is available for this data, an initial training step is performed on $\mathcal{D}_s$ to produce $\Theta_F$, using the same architecture described above. This is followed by the pruning step, producing $\mathcal{D}_{ss}$. Next, to tune hyperparameters, $\mathcal{D}_{ss}$ is randomly split into training and validation subsets, at an 80-20 ratio. We can see then that, advantageously, pruning can be exploited to perform adaptive hyperparameter tuning as well, as the validation set will also be aligned with the test distribution. A random search of size 10 is used for hyperparameters (see [1] for details); in particular, the pruning ratio $N_{ss}/N_s$ is drawn from a uniform distribution between 20% and 99%. The entire set of experiments is repeated 3 times for reproducibility, using different random seeds for hyperparameters, weight initialisations, and dataset splits. All other hyperparameter choices and training details follow the DomainBed default options.

In total, we compare 8 combinations of training strategies. This includes 2 baseline training algorithms used to optimise $\Theta$: ERM, and the adaptive CORAL [28] algorithm, which also exploits $\mathcal{D}_t$. We then enhance these with either AdaPrune, KMM, or landmark selection. We report the results in Table 1, along with standard errors across the 3 repeats.

When training with ERM, AdaPrune provides an average accuracy gain of around 4%, outperforming the 3% gain of KMM. On its own, this is less than the 10% improvement of training the model with CORAL. However, interestingly, we can see that the adaptative effects of AdaPrune and CORAL are complementary to each other, meaning that combining the two gives the highest performance out of all the methods we test, a 14% improvement over non-adaptive training. On the other hand, KMM combined with CORAL, along with landmark selection, had a substantially negative impact on accuracy.

### A. t-SNE Analysis

In this section, we wish to better understand how examples are being pruned from the training dataset. We do this by generating t-SNE [29] plots of the feature embeddings $z_i^s$ and $z_j^t$ (Figure 3). For the specific training scenarios shown, we also report $\text{MMD}(\mathcal{D}_s, \mathcal{D}_t)$, $\text{MMD}(\mathcal{D}_{ss}, \mathcal{D}_t)$, accuracy of the model trained on $\mathcal{D}_s$, accuracy when trained on $\mathcal{D}_{ss}$, and the proportion of training examples which have been removed.

The plots clearly show that the pruned training instances are the ones furthest from the target examples, which is a good indicator that these instances will have the least relevance to the test data. Removing these corresponds to a decrease in MMD and an increase in test-domain accuracy.

### B. MMD vs Accuracy Analysis

Finally, we analyse the relation between the MMD between the training and test domains, and model accuracy on the test domain. A negative correlation between these two variables would substantiate reducing the MMD as a valid objective for data pruning. Figure 4 shows a scatter plot of MMD vs accuracy for the baseline ERM models for all repeats and hyperparameter sweeps (120 datapoints in total), along with the Pearson correlation and its associated p-value. The plot clearly shows the expected negative trend: that is, models trained on datasets more similar to the test data tend to perform better on that test data, with the value of $r = -0.56$ indicating a moderate correlation between the two variables. This evidence confirms the use of the MMD as an effective criterion for pruning training data.

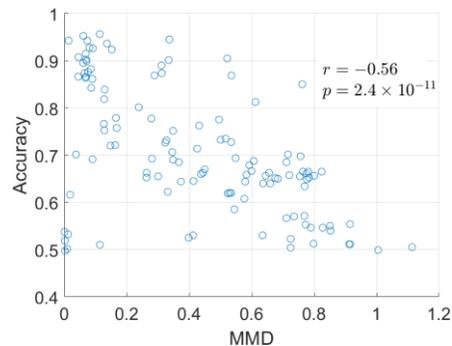

Figure 4: Scatter plot of the MMD between the training and test domains vs model accuracy on the test domain.

### III. CONCLUSION

This paper introduced AdaPrune, a novel method for UDA based on removing instances from the training dataset. This way, only the examples most relevant to the test domain are used to train the model. We showed that AdaPrune outperforms its most closely related alternatives, and is complementary to other UDA approaches such as CORAL. Additionally, the optimisation procedure makes reasonable choices in determining the most irrelevant training examples to be removed, and its objective function is well-correlated with test domain accuracy.

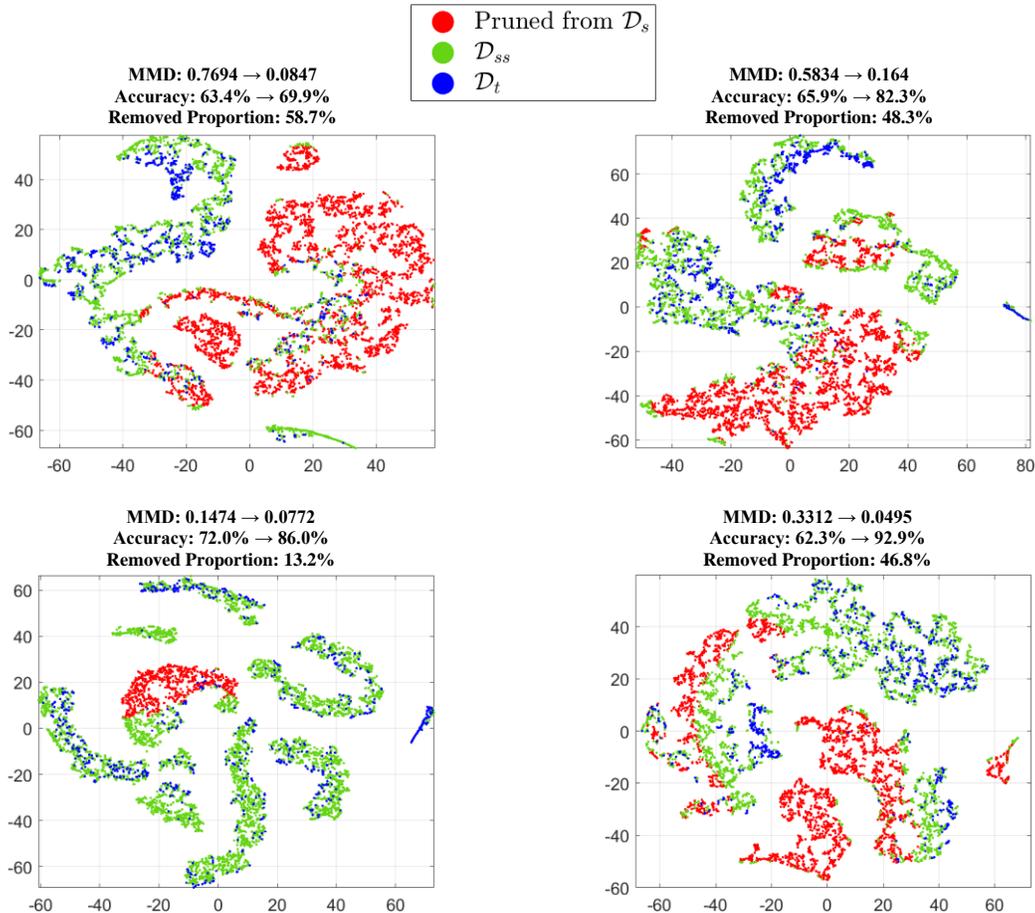

Figure 3: t-SNE plots of source and target feature embeddings for 4 exemplar training scenarios. Red points indicate training examples which have been pruned from $\mathcal{D}_s$, green points are the training examples which have been retained, and blue points are the target examples $\mathcal{D}_t$. Also reported are the MMD and model accuracy before and after pruning, and the proportion of training examples removed.

As with KMM, AdaPrune has cubic complexity in the size of the training set, but can be easily and effectively scaled in the same manner as KMM (e.g., via bootstrap aggregation [30]). Further investigation into scaling this method for larger datasets would form a good basis for future work.

## IV. ACKNOWLEDGEMENTS

This work was supported by grants from BAE Systems and the Engineering and Physical Sciences Research Council. The authors acknowledge the use of the IRIDIS High Performance Computing Facility, and associated support services at the University of Southampton, in the completion of this work.